\documentclass{article} 
\usepackage{iclr2023_conference,times}

\usepackage{amsmath,amsfonts,bm}

\def\eqref#1{equation~\ref{#1}}

\def\1{\bm{1}}

\DeclareMathAlphabet{\mathsfit}{\encodingdefault}{\sfdefault}{m}{sl}
\SetMathAlphabet{\mathsfit}{bold}{\encodingdefault}{\sfdefault}{bx}{n}

\usepackage{hyperref}
\usepackage{paralist}
\usepackage{amsmath}
\usepackage{amssymb}
\usepackage{wrapfig}
\usepackage{multirow}
\usepackage{multicol}
\usepackage{multirow}

\usepackage{dsfont}
\usepackage{tabularx}
\usepackage{graphicx}
\usepackage{xcolor}
\usepackage{soul}
\usepackage{xspace}
\usepackage{booktabs}
\usepackage{caption}
\usepackage{enumitem}
\usepackage{algorithm}
\usepackage[noend]{algorithmic}
\usepackage{pythonhighlight}
\usepackage{pifont}

\usepackage[most]{tcolorbox}
\usepackage[framemethod=tikz]{mdframed}
\definecolor{qualcolor}{RGB}{128,64,0}
\gdef\Sepline{%
  \par\noindent\makebox[\linewidth][l]{%
  \hspace*{-\mdflength{innerleftmargin}}%
   \tikz\draw[thick,dashed,gray!60] (0,0) --%
        (\textwidth+\the\mdflength{innerleftmargin}+\the\mdflength{innerrightmargin},0);
  }\par\nobreak}
  
\newcommand{\cmark}{\ding{51}}%
\newcommand{\xmark}{\ding{55}}%

\usepackage[normalem]{ulem}
\usepackage[colorinlistoftodos]{todonotes}
\definecolor{green(pigment)}{rgb}{0.0, 0.65, 0.31}
\usepackage{array}
\usepackage{makecell}

\usepackage{framed}
\usepackage{hhline}

\usepackage{courierten}
\usepackage[T1]{fontenc}    
\newcommand\modelfont[1]{{\usefont{T1}{courierten}{m}{n}#1}}

\newcommand{\methodnamelong}{\modelfont{Self-Correction for Sequence Generation}\xspace}

\newcommand{\methodnameshort}{\modelfont{Self-Correction}\xspace}

\newcommand{\method}{\textsc{Self-Correct}\xspace}
\newcommand{\myparagraph}[1]{\par\noindent\textbf{{#1}}} 

\title{Generating Sequences by \\Learning to [Self-]Correct}

\author{Sean Welleck\textsuperscript{1,3,*} \hspace{1pt}
Ximing Lu\textsuperscript{1,*}
\AND Peter West\textsuperscript{3,$\dagger$} \hspace{1pt} Faeze Brahman\textsuperscript{1,$\dagger$}
\AND Tianxiao Shen\textsuperscript{3} \hspace{1pt} Daniel Khashabi\textsuperscript{2} \hspace{1pt} Yejin Choi\textsuperscript{1,3} \\
\textsuperscript{1}Allen Institute for Artificial Intelligence\\
 \textsuperscript{2}Center for Language and Speech Processing,  Johns Hopkins University \\
  \textsuperscript{3}Paul G. Allen School of Computer Science \& Engineering, University of Washington \\
}

\iclrfinalcopy 
\begin{document}

\maketitle
\renewcommand*\thefootnote{\textbf{$*$}}\footnotetext{First authors, contributed equally. \textbf{$\dagger$}Second authors, contributed equally.}

\renewcommand*{\thefootnote}{\arabic{footnote}}
\setcounter{footnote}{0}
\begin{abstract}
Sequence generation applications require satisfying semantic constraints, such as ensuring that programs are correct,
using certain keywords, 
or avoiding undesirable content. 
Language models, whether fine-tuned or prompted with few-shot demonstrations, frequently violate these constraints, and lack a mechanism to iteratively revise their outputs.
Moreover, some powerful language models are  of extreme scale or inaccessible, making it inefficient, if not  infeasible, to update their parameters for task-specific adaptation. 
We present \textsc{self-correction}, 
an approach that decouples an imperfect base generator (an off-the-shelf language model or supervised sequence-to-sequence model) from a separate corrector that learns to iteratively correct  imperfect generations.
To train the corrector, we propose an online training procedure that can use either scalar or natural language feedback on intermediate imperfect generations. 
We show that  \textsc{self-correction} improves upon the base generator in three diverse generation tasks-- mathematical program synthesis, lexically-constrained generation, and toxicity control-- even
when the corrector is much smaller than the base generator.

\end{abstract}

\section{Introduction}
The standard practice for natural language generation tasks is inherently single-pass: applying a decoding procedure 
to either a few-shot prompted language model or one tuned for a given task, then considering the generation as ``finished''~(e.g. \citet{radford2019language,brown2020,chen2021codex}).
Powerful generation models
often meet most of the task requirements, yet miss a few
(e.g., omitting a subset of keywords), 
or generate incorrect hypotheses that nevertheless provide useful structure 
(e.g., a correct problem solving strategy with a missing step). 
However, after generating even a slightly sub-optimal sequence, the single-pass paradigm requires models to ``start from scratch'',  
effectively discarding work already done. 
A more natural, intuitive approach
is leveraging the generation as a useful starting point
to refine 
into a higher quality output.

\begin{figure}[t]
    \centering
    \includegraphics[width=0.99\textwidth]{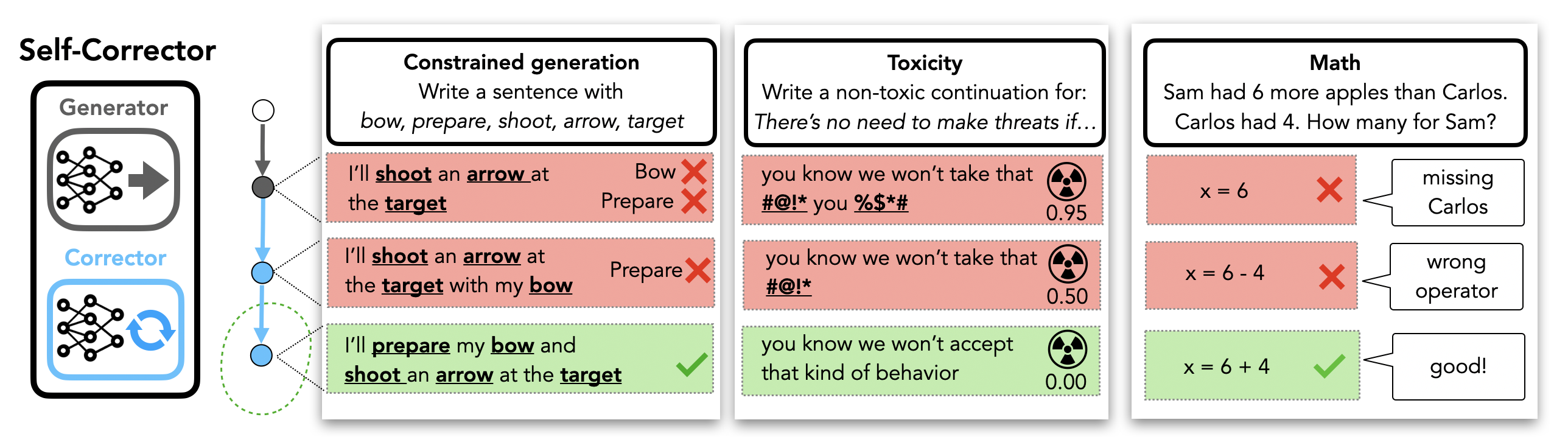}
    \caption{\textsc{Self-corrector}s decompose generation into a base generator that proposes an initial hypothesis, and a corrector that iteratively improves its quality.}
    \label{fig:teaser}
\end{figure}

\begin{figure}[t]
    \centering

    \includegraphics[width=0.99\textwidth]{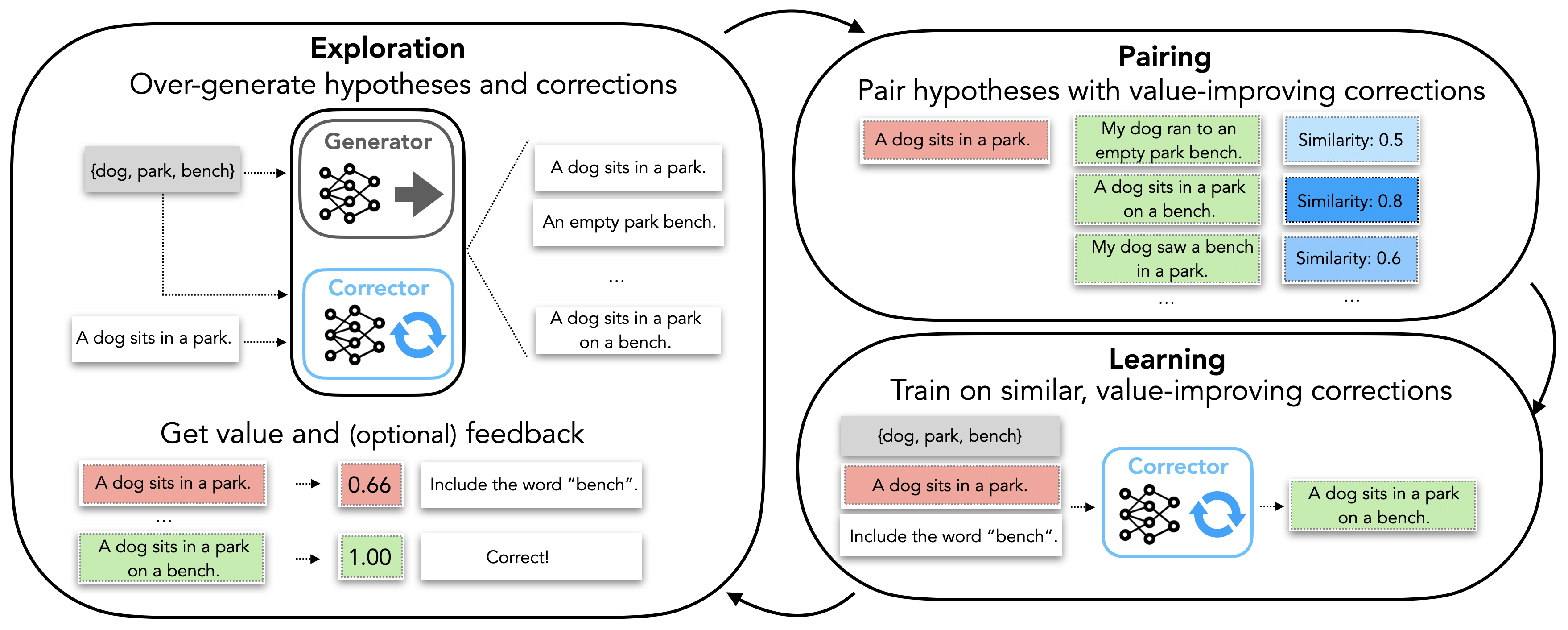}
    \caption{\textsc{Self-corrective learning} iteratively trains a corrector by generating hypotheses and corrections, forming value-improving pairs, and selecting those with high similarity for learning.
    }
    \label{fig:learning}
\end{figure}

To formalize this intuition, we introduce \methodnamelong. 
\autoref{fig:teaser} demonstrates its central principle: a generation model is re-framed as a base
 \emph{generator}, which produces a reasonable initial hypothesis but does not need to solve the task in one pass, and a second module--the \emph{corrector}--trained to 
make up the difference between the hypothesis and an optimal solution. Neither the generator nor the corrector must solve the full task in one pass, and the corrector can be applied multiple times
to iteratively improve the output (\S\ref{subsec:ablation}). We propose a simple, general procedure for training the corrector (\autoref{fig:learning}) by pairing generator outputs with carefully selected targets. 
The result is a system which self-corrects, producing outputs through multiple generation passes and breaking the task into steps that can be solved by dedicated and efficient sub-systems.

We find that \methodnameshort is broadly applicable. Training a corrector model improves the base generator on 3 diverse tasks: mathematical program synthesis (\S\ref{ssec:math}), lexically constrained generation (\S\ref{ssec:constrained}), and toxicity reduction (\S\ref{ssec:toxicity}). The trained corrector model can even be applied to a larger generator with similar performance to training a new corrector (\S\ref{sec:modularity}), showing that the sub-task of correction is transferable, even to stronger generators. Finally, we explore the prospect of introducing a third module to the \methodnameshort system (\S\ref{sec:feedback})--explicitly using natural language feedback to guide corrections--with promising results. \methodnameshort offers an exciting opportunity to build on existing generation models and the sequences they generate, with efficient, effective, and transferable corrector networks.

\section{Self-correcting sequence generators}
\label{sec:method}

A typical autoregressive text generator (e.g. GPT-3~\citep{brown2020}) maps an input prompt to a distribution over outputs using a single parameterized module (e.g. a large transformer), $p_0(y|x)$.
We explore an alternative that decomposes into two modules, a base \textit{generator}, and a \textit{corrector},
\begin{align}
\label{eqn:model-onestep}
p(y|x)=\sum_{y_0}\underbrace{p_0(y_0|x)}_{\text{generator}}\underbrace{p_\theta(y|y_0,x)}_{\text{corrector}}
\end{align}
where the generator provides an initial hypothesis that is refined by the corrector.
In practice, the corrector can be applied multiple times, $p(y_T|x)=\sum_{y_0}\sum_{y_1}\cdots \sum_{y_{T-1}}p_0(y_0|x)\prod_t p_\theta(y_{t+1}|y_t,x)$.
Since a model of this form can both generate and correct its generations, we call it a \modelfont{Self-Corrector}.

Self-correctors have several unique properties compared to typical generators.
First, a self-corrector 
decouples generation and correction, 
allowing us to \emph{freely parameterize each module} -- 
for instance, by prompting a single language model or using two different language models.  
In this paper, we develop a framework to train a separate corrector model
(\S\ref{ssec:learning}).
We find that the resulting self-corrector improves upon the generator alone (\S\ref{sec:exprs}), even when the corrector is much smaller (\S\ref{sec:modularity}).

Second, since the generator and the corrector are separated, we can keep the generator as a general-purpose language model and \emph{train the corrector with different objectives} for different task requirements.
In \S\ref{ssec:learning}, we propose a training algorithm for the corrector that is dedicated to improving generations, where the improvement can be in any aspect, measured by scalar values.

Third, the corrector can receive \textit{explicit feedback} about intermediate generations to guide subsequent generations.
Formally, $p(y|x)=\sum_{y_0}p_0(y_0|x)p_\theta(y|y_0,x,f(y_0))$, where $f$ is the feedback.
The feedback can be of many forms, e.g. a sentence, a compiler trace, etc. 
In contrast, a typical generator that generates in a single pass does not leverage feedback on its own generation.
In this paper, 
we show that the corrector can learn to exploit explicit natural language feedback to achieve better performance~(\S\ref{sec:feedback}).
Next, we describe our training framework of the corrector.

\subsection{Learning a Corrector}
\label{ssec:learning}
Our goal is to have the generator generate an initial hypothesis, 
then improve the hypothesis with the corrector (Eq.~\ref{eqn:model-onestep}).
We train the corrector to improve the quality of a hypothesis, while staying as close as possible to the original hypothesis. 
Here, quality is measured with a scalar value function $v(y)$ which we assume is accessible at training time (e.g. a classifier).

Since direct supervision on how to improve hypotheses is not available, we design a new algorithm to train the corrector, which we refer to as self-corrective learning.
The algorithm collects a pool of generations, 
groups them and 
 selects pairs of generation 
that increase in value and are nearby, then updates the corrector on these pairs.
As training progresses, more 
generations are added to the pool using the current corrector.
Algorithm~\ref{alg:main} summarizes self-corrective learning, detailed below.

\myparagraph{Initialization.}
Self-corrective learning begins with a generator $p_0(y_0|x)$, a corrector 
$p_\theta(y'|y,x)$
, a set of training prompts $X$, and a value function 
$v:\mathcal Y \rightarrow\mathbb R$.
Optionally,
we can use additional feedback 
$f: \mathcal Y \rightarrow \mathcal F$ and learn $p_\theta(y'|y,x,f(y))$,
where $\mathcal F$ 
is arbitrary.

The algorithm initializes a datapool of (input, output, value, feedback) examples by using the generator to generate multiple outputs for each input.
Formally, 
\begin{align}
D_x=\{(x,y,v(y), f(y))\ |\ \text{for all } y\in y^{1:N}\sim q(p_0(\cdot|x))\},\quad D=\bigcup_{x\in X} D_x,
\label{eqn:d0}
\end{align}
where $y^{1:N}$ denotes $N$ outputs generated with decoding algorithm $q$ (e.g. temperature sampling).
When available, $(x, y, v(y), f(y))$ examples from another source (e.g. a dataset) can also be added.

\newcommand{\algcomment}[1]{{\footnotesize \fontfamily{cmtt}\selectfont // #1}}
\renewcommand{\algorithmiccomment}[1]{\hfill{\(\triangleright\)~#1}\par}
\begin{figure}[t]
\vspace{-1em}
\begin{algorithm}[H]
\small
\begin{algorithmic}
\INPUT{Generator $p_0$, corrector $p_\theta$, prompts $X$, value $v(\cdot)$, feedback $f(\cdot)$}
\\
\text{Initialize datapool }$D$ by sampling from $p_0$\algorithmiccomment{Initialization: Eq.~\ref{eqn:d0}}

\FOR{$\text{iteration}\in\{1,2,\ldots\}$}
\FOR{$x \in X$} 
\STATE Sample hypotheses $y$ from datapool $D$  
\STATE Generate corrections $y'\sim p_\theta(\cdot|y,x,f(y))$
\STATE Add all $(x,y',v(y'),f(y'))$ to the datapool $D$
\algorithmiccomment{Exploration: Eq.~\ref{eqn:exploration}}

\ENDFOR
Form value-improving pairs $P$ from $D$\algorithmiccomment{Pairing: Eq.~\ref{eqn:pairing}}
\FOR{step in $1,2,\ldots,M$ }
\item Sample a batch of value-improving pairs from $P$ using Eq.~\ref{eqn:residual-subsample}
\item Compute the loss and update $\theta$ using gradient descent
\algorithmiccomment{Learning}
\ENDFOR
\ENDFOR
\end{algorithmic}
\caption{Self-corrective learning}
\label{alg:main}
\end{algorithm}
\vspace{-1em}
\end{figure}

\myparagraph{Pairing.}
Next, self-corrective learning forms \textit{value-improving pairs}: examples of mapping a hypothesis to a higher-valued correction.
We use 
the  datapool $D$ to form a set of (input, hypothesis, correction) pairs. 
A pair is formed when an output has a higher value than another
\footnote{We also store the value and feedback for $y$ and $y'$ along with $(x,y,y')$, which we omit to reduce clutter.}:
\begin{align}
\label{eqn:pairing}
    P_x=\{(x,y,y')\mid v(y)<v(y')\text{ for all } y,y'\in D_x\times D_x\},\quad P=\bigcup_{x\in X} P_x,
\end{align}

\myparagraph{Learning.}
Next, self-corrective learning selects (input, hypothesis, correction) pairs to update the corrector with.
We sample a $(x,y,y')$ pair proportional to its improvement in value as well as the proximity between the hypothesis $y$ and the correction $y'$:
\begin{align}
\label{eqn:residual-subsample}
  \mathds P[(x, y,y')]&\propto \exp\big(\underbrace{\alpha\cdot(v(y')-v(y))}_{\text{improvement}}+\underbrace{\beta\cdot s(y,y')}_{\text{proximity}}\big)/Z(y), 
\end{align}
where $s(y,y')$ is a similarity function and $Z(y)$
normalizes over the available corrections for $y$ in $P_x$.
Increasing the hyperparameter $\alpha\in\mathbb{R}_{\geq 0}$ puts more weight on targets that add more value, while
increasing  $\beta\in\mathbb{R}_{\geq 0}$ retains more similar targets. 
We update the corrector using the cross-entropy loss $\mathcal{L}(\theta) = -\log p_\theta(y'|y,x,f(y))$ on batches sampled in this way.

\myparagraph{Exploration.}
During exploration, self-corrective learning adds 
new generations to the datapool by generating from the current corrector:
\begin{align}
\label{eqn:exploration}
D'_x&=\{(x,y',v(y'), f(y'))\ |\ \text{for all } y'\in y'^{1:N}\sim q(p_\theta(\cdot|y,x,f(y))\},\quad D'=\bigcup_{x\in X} D'_x
\end{align}
and updating the datapool $D\leftarrow D\cup D'$.
The hypotheses $y$ to correct can come from any source, e.g. newly sampled from the base generator,
or from the datapool; 
we use the latter in our experiments.

\myparagraph{Inference.}
We use the trained corrector along with a generator to generate a trajectory $y_0,y_1,\ldots,y_T$, and consider $y_T$ the final output.
Since marginalizing over the intermediate generations in Eq.~\ref{eqn:model-onestep} is intractable, we approximate each summation with a single sequence generated with a decoding algorithm $q(\cdot)$.
That is, we decode from the generator, then repeatedly from the corrector:
\begin{itemize}[leftmargin=*,topsep=0pt,itemsep=-1ex,partopsep=1ex,parsep=1ex]
    \item Generation: $y_0\sim q(p_0(y_0|x))$;
    \item Correction: $y_{t+1}\sim q(p_\theta(y_{t+1}|y_{t},x, f(y_t)))$,\quad $t=0,1,\dots,T-1$.
\end{itemize}
The stopping time $T$ is either fixed, or when a target value is obtained (if $v(y)$ is available).

\section{Experiments}
\label{sec:exprs}
We evaluate \textsc{self-correction} on a diversity of tasks: \textbf{mathematical program synthesis}, in which generations are strictly correct or incorrect, 
and generators typically have low performance; 
\textbf{lexically-constrained generation}, which allows for partial credit, and generators usually give partially-correct solutions (e.g. matching 3 out of 5 constraints); and \textbf{toxicity control}, 
where `correctness' is more loosely defined, 
and the output space is much more open-ended.
Our experiments are organized to study three settings:
\begin{enumerate}[leftmargin=*,topsep=0pt,itemsep=-1ex,partopsep=1ex,parsep=1ex]
\item  Using self-correctors to improve upon generators (\S\ref{ssec:math},\ref{ssec:constrained},\ref{ssec:toxicity}).
\item Correcting generators that are much larger than the corrector (\S\ref{sec:modularity}).
\item Leveraging explicit feedback during training and inference (\S\ref{sec:feedback}).
\end{enumerate}
Next, we describe the self-correction setup and baselines for each task, along with their results.  \footnote{Code will be publicly available upon acceptance.}

\subsection{Mathematical Program Synthesis}
\label{ssec:math}
First, we consider mathematical program synthesis~\citep{austin2021ProgramSW,mishra2022lila}.
Given a natural language problem specification $x$, the task is to generate a  program $y$ that upon execution returns the correct answer to $x$.
The task is challenging as it draws on language understanding, multiple-step mathematical problem solving (e.g. identifying a solution strategy, decomposing a problem), and leveraging symbolic tools (e.g. built-in operations, variables).
Furthermore, the task demands a high level of precision, e.g. a single misplaced operation  makes the program incorrect.

\myparagraph{Experimental setup.}
As the corrector we use GPT-Neo 1.3B~\citep{gpt-neo}, an open-source autoregressive language model.
GPT-Neo is pre-trained on language and code~\citep{pile}, and hence is widely used 
for code-related generation (e.g. \citet{chen2021codex,ni2022learning,mishra2022lila}).
We consider two settings for the initial generator: (1) a separate fine-tuned instance of GPT-Neo 1.3B, and (2) few-shot prompted GPT-3~\citep{brown2020}.
For GPT-3, we evaluate the davinci and text-davinci-002 engines, representative of large ($\approx 175B$\footnote{Estimated size of \textit{davinci} ({\scriptsize \url{https://blog.eleuther.ai/gpt3-model-sizes}}). Further details not available.}) generators that are state-of-the-art in related tasks~\citep{Wei2022ChainOT}.
See the Appendix for additional details.

\myparagraph{Self-correction setup.} As the value function we use correctness, which is 1 when the program $y$ executes and outputs the ground-truth answer and 0 otherwise.
Our main experiments do not use explicit feedback, i.e. $f(y)=\emptyset$.
At inference time, we study two settings for the corrector: (1) applying $k$ corrections and selecting the final generation, (2) an oracle  setting that only corrects a draft if the draft is incorrect.
We use greedy decoding for the generator and corrector, and $k=1$.

\begin{table}[t]
\begin{minipage}[t]{0.45\linewidth}
\centering\footnotesize
\renewcommand\arraystretch{1.14}
   \begin{tabular}{clcc}
    \toprule
         \textbf{Dataset} & \textbf{Model} & \textbf{Correct}\\
         \midrule
         \textbf{Multiarith} & GPT-NEO 1.3B & 60.00\\
         & +\textsc{Self-Correct} & \textbf{98.33}\\
         & +$\textsc{Self-Correct}_*$ & \textbf{99.17}\\
         \midrule
         \textbf{Multitask} 
         & GPT-NEO  1.3B  & 49.02\\
         & +\textsc{Self-Correct}& \textbf{73.53}\\
         & +$\textsc{Self-Correct}_*$ & \textbf{78.24}\\
    \bottomrule
    \end{tabular}
\end{minipage}
\hfill
\begin{minipage}[t]{0.54\linewidth}
\centering \footnotesize
     \begin{tabular}{clcc}
    \toprule
         \textbf{Dataset} & \textbf{Model} & \textbf{Params} & \textbf{Correct}\\
         \midrule
         \textbf{GSM} 
         & \textit{OpenAI 3B}~[\citenum{cobbe2021gsm8k}] & 3B& 15.50\\
         & \textit{OpenAI 6B}~[\citenum{cobbe2021gsm8k}] & 6B\citenum{}& 20.00\\
         & GPT-NEO~[\citenum{ni2022learning}] & 2.7B\citenum{}& 18.80\\
         & NEO FCP+PCP~[\citenum{ni2022learning}] & 2.7B\citenum{}& 19.50\\
         \cmidrule(lr){2-4}
         & GPT-NEO & 1.3B & 8.57\\
         & +\textsc{Self-Correct} & 1.3B & \textbf{21.26}\\
         & +$\textsc{Self-Correct}_*$ & 1.3B & \textbf{24.22}\\
    \bottomrule
    \end{tabular}
\end{minipage}
\caption{Evaluation results of mathematical program synthesis experiments. GPT-NEO (1.3B) is the initial generator for \textsc{Self-Correct}. 
    $\textsc{Self-Correct}_*$ means only applying the corrector to incorrect outputs.
    \textit{Italicized}: original non-program version of GSM. 
    }
    \label{tab:math_results}
\end{table}

\setlength{\columnsep}{0.2cm}
\begin{figure}[t]
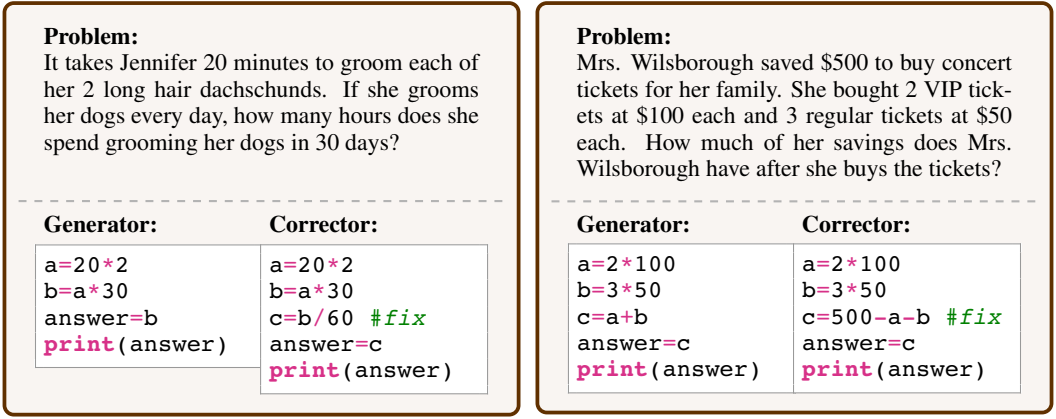

\begin{multicols}{2}

\vspace{2mm}

\begin{tcolorbox}[colback=qualcolor!5!white,colframe=qualcolor!75!black]
\begin{small}
\textbf{Problem:}\\
It takes Jennifer 20 minutes to groom each of her 2 long hair dachschunds. If she grooms her dogs every day, how many hours does she spend grooming her dogs in 30 days?
\vspace{1.1em}
\Sepline
\vspace{-1.2em}
\begin{multicols}{2}
\textbf{Generator:}
\vspace{-0.3em}
\begin{python}
a=20*2
b=a*30
answer=b
print(answer)
\end{python}
\vspace{1.1em}
\columnbreak

\textbf{Corrector:}
\vspace{-0.3em}
\begin{python}
a=20*2
b=a*30
c=b/60 #fix
answer=c
print(answer)
\end{python}
\end{multicols}
\end{small}
\vspace{-2em}
\end{tcolorbox}

\begin{tcolorbox}[colback=qualcolor!5!white,colframe=qualcolor!75!black]
\begin{small}
\textbf{Problem:}\\
Mrs. Wilsborough saved \$500 to buy concert tickets for her family. She bought 2 VIP tickets at \$100 each and 3 regular tickets at \$50 each. How much of her savings does Mrs. Wilsborough have after she buys the tickets?
\Sepline
\vspace{-1.2em}
\begin{multicols}{2}
\textbf{Generator:}
\vspace{-0.3em}
\begin{python}
a=2*100
b=3*50
c=a+b
answer=c
print(answer)
\end{python}
\columnbreak

\textbf{Corrector:}
\vspace{-0.3em}
\begin{python}
a=2*100
b=3*50
c=500-a-b #fix
answer=c
print(answer)
\end{python}
\end{multicols}
\end{small}
\vspace{-2em}
\end{tcolorbox}
\end{multicols}
\vspace{-1.7em}
\caption{\textbf{Grade-school-math (GSM) self-corrections.} On the left, the corrector fixes the units (from minutes to hours) in the generator's solution. On the right, the corrector revises the logic so that the program computes the total savings instead of the spent on tickets. We add \textit{\#fix} here to indicate the change.
See \autoref{fig:math-examples2} and \autoref{fig:math-examples3} for additional examples.\small}
\label{fig:examples}
\end{figure}

\myparagraph{Datasets.} We evaluate on problems from 5 problem solving datasets: MultiArith~\citep{roy2015multiarith}, AddSub~\citep{hosseini2014addsub}, SingleOp~\citep{roy2015multiarith}, SVAMP~\citep{patel2021svamp}, and GSM8k~\citep{cobbe2021gsm8k}.
As in prior work \citep{austin2021ProgramSW,ni2022learning,mishra2022lila}, we frame these as program synthesis by converting their solutions to Python programs.\footnote{{We use data from the Lila benchmark ({\scriptsize \url{https://github.com/allenai/Lila}).}}}
We separate our experiments into three increasingly difficult settings: 
\begin{enumerate}[leftmargin=*,topsep=0pt,itemsep=-1ex,partopsep=1ex,parsep=1ex]
    \item \textbf{MultiArith}, using problems from the MultiArith arithmetic word problem dataset.
    \item \textbf{Multitask}, using problems from 4  arithmetic datasets (MultiArith,  AddSub, SingleOp, SVAMP).
    \item \textbf{GSM}, using problems from the challenging GSM8k dataset.
\end{enumerate} 
For the MultiArith and Multitask settings, we make train/valid/test splits using 60/20/20\% of the respective datasets.
Similar to \citet{ni2022learning}, for the GSM setting we use the official GSM8k test split, and create a validation split using 20\% of the training set.
Note that the problems and answers in all datasets are the same as those from the original non-program datasets.

\myparagraph{Baselines.} We compare \textsc{self-correct} with its baseline generator (GPT-Neo 1.3B) in all three settings. For the GSM setting, we compare with existing work that uses models within the same magnitude of scale, including NEO FCP+PCP~\citep{ni2022learning}, which tunes GPT-NEO 2.7B with additional self-sampled programs, and their fine-tuned GPT-NEO 2.7B baseline.
We also report 3B and 6B fine-tuned GPT3-like language models from \citet{cobbe2021gsm8k}, which were trained on the non-program version of GSM8k.
We evaluate larger models later in (\S\ref{sec:modularity}).

\myparagraph{Results.} 
As seen in \autoref{tab:math_results},
the self-corrector improves upon the generator in all three settings, using either inference strategy:  always correcting  (\textsc{self-correct}), or only correcting incorrect solutions  (\textsc{self-correct}$_*$).
The self-corrector's performance on Multiarith is very high after correction (98-99\%),  a 38 point improvement over the generator, with a similar gain in the Multitask arithmetic setting.
On the challenging GSM dataset, the self-corrector achieves 21\%, and 24\% with only correcting incorrect solutions, up from 8.57\% for the generator.
Notably, this is higher than previous work based on the larger 2.7B GPT-Neo, 
or larger models tuned on the language version of GSM.
The results show that self-corrective learning can improve task performance via training a corrector.
Qualitatively, the self-corrector can correct values in a correctly structured solution, fix the order of operations within a multistep solution, adjust unit conversions, and make larger multipart revisions (see Figures~\ref{fig:examples},\ref{fig:math-examples2},\ref{fig:math-examples3}). Notably, these are learned automatically through self-corrective learning.

\subsection{Lexically Constrained Generation}
\label{ssec:constrained}
Next, we consider lexically constrained generation. Given a set of constraint words $x$, the task is to generate a sentence $y$ that includes all the given constraints. 
Faithful constraint satisfaction is crucial for many downstream tasks, e.g., those that require converting information to text~\citep{mckeown_1985}.

\myparagraph{Datasets and Metrics.} We experiment on \textsc{CommonGen}~\citep{lin-etal-2020-commongen} and E2E~\citep{novikova-etal-2017-e2e}. \textsc{CommonGen} is a benchmark for generative commonsense reasoning where the task is to generate a coherent sentence given a set of words (e.g., dog, catch). 
E2E involves converting structured inputs into natural language.
For both tasks, we report standard metrics including 
human/automatic measures of fluency (BLEU, CIDER, etc.) as well as constraint coverage. We collect human measures of fluency on Amazon Mechanical Turk; see the Appendix for  details.

\myparagraph{Setup.} We parameterize the base generator with GPT-2 \cite{radford2019language} (large-size for \textsc{CommonGen} and medium-size for E2E).  
We fine-tuned the generator for each task. 
As the value function for self-corrective learning we use coverage, i.e. the percentage of constraints that are present in the output.
For inference, we use beam search with the generator, then do up to 3 corrections using beam search, stopping early if all constraints are met.
See the Appendix for additional details.

\begin{table}[t]
\begin{minipage}[t]{0.54\linewidth}
\centering\footnotesize
\setlength{\tabcolsep}{4pt}
   \begin{tabular}{lccc}
    \toprule
         \textbf{Method} & \textbf{Runtime} & \textbf{CIDER} & \textbf{Constraints} \\
         \midrule
         NeuroLogic~[\citenum{lu-etal-2021-neurologic}] &  2.04s & 14.70 & 97.70\\
         NeuroLogic-A*~[\citenum{lu2022neurologicastar}]& 19.24s & 15.20 & 97.80 \\
         \midrule
         GPT-2 & 0.20s & 14.97 & 91.38 \\
         \method & 0.80s& 15.30 & 94.58 \\
         \ \ +NeuroLogic & 2.24s& 15.28 & \textbf{97.80} \\
    \bottomrule
    \end{tabular}
\end{minipage}
\hfill
\begin{minipage}[t]{0.45\linewidth}
\centering \footnotesize
\renewcommand\arraystretch{1.2}
\scalebox{.86}{
     \begin{tabular}{lcc}
    \toprule
         \textbf{Method} & \textbf{Fluency} & \textbf{Constraints} \\
         \midrule
         Prefix-Tuning~[\citenum{li-liang-2021-prefix}] & 2.96 & 91.16 \\
          NeuroLogic~[\citenum{lu-etal-2021-neurologic}] & 2.80 &  96.91\\
          NeuroLogic-A*~[\citenum{lu2022neurologicastar}] & 2.85 &  96.97\\
         \midrule
         GPT-2 & 2.94 & 91.50 \\
         \method& \textbf{2.98} & \textbf{98.77}\\
    \bottomrule
    \end{tabular}}
\end{minipage}
\caption{\textbf{Lexically-constrained generation.} By training a corrector to optimize constraint satisfaction, \method improves constraints while maintaining fluency, without modifying the underlying generator. Due to space, we show CIDER for \textsc{CommonGen} and human judgement for E2E as measures of fluency. Other metrics show similar trends and can be found in the Appendix.
}
    \label{tab:lexical_results}
\end{table}

\myparagraph{Results.} Table~\ref{tab:lexical_results} shows the evaluation results. 
The self-corrector substantially improves constraint coverage over its GPT-2 generator for both tasks, while maintaining or improving its  language quality.
On the \textsc{CommonGen} benchmark, the self-corrector paired with the NeuroLogic constrained decoding algorithm ~\citep{lu-etal-2021-neurologic} achieves the best results, outperforming the more sophisticated NeuroLogic-A* decoding algorithm, while being an order of magnitude faster. 
Notably, on E2E, self-correction \textit{outperforms} Neurologic-A* decoding, despite only using standard beam search. 
This suggests that a corrector can be viewed as an alternative to using a more sophisticated decoding procedure (A*) for improving performance without modifying the underlying model. See \autoref{fig:examples-cg-e2e} for qualitative examples.

\subsection{Toxicity Reduction}
\label{ssec:toxicity}
Next, we consider the task of toxicity reduction~\citep{gehman-etal-2020-realtoxicityprompts,liu-etal-2021-dexperts}.
Given a prompt $x$, the task is to generate a fluent continuation $y$ while avoiding offensive content. 
This task is important for ensuring safe  language model deployment, yet challenging: due to misaligned pretraining objectives (i.e. modeling internet text vs. non-toxic text), 
language models are susceptible to generating toxic completions, even when prompted with seemingly innocuous text~\citep{gehman-etal-2020-realtoxicityprompts}.
Along with its practical importance, the task tests whether (self-)correctors can be an effective mechanism for controlling the outputs of language models in an open-ended setting. 

\myparagraph{Datasets and Metrics.} We use the \textsc{RealToxicityPrompts} benchmark~\citep{gehman-etal-2020-realtoxicityprompts} which contains 100k prompts designed to elicit toxic generations. Following the experimental setup of~\citet{liu-etal-2021-dexperts}, during training we use 85K prompts from the training set, and for evaluation we use the same 10K non-toxic prompts from test set as \citet{liu-etal-2021-dexperts}.
We use Perspective API to measure \textit{maximum toxicity}, defined as the average maximum toxicity over 25 sampled generations, and the (empirical) \textit{toxicity probability} of at least 1 out of 25 generations being toxic.

\myparagraph{Baselines.}
We compare \method with its generator (GPT-2) and previously reported baselines from~\citet{quark22}, including PPLM~\citep{Dathathri2020PlugAP}, GeDi~\citep{krause-etal-2021-gedi-generative}, DExpert~\citep{liu-etal-2020-unsupervised}, DAPT~\citep{gururangan-etal-2020-dont}, PPO~\citep{quark22}, and Quark~\citep{quark22}.
The latter two -- Proximal Policy Optimization (PPO) and Quantized Reward Konditioning (Quark) -- represent strong, state-of-the art approaches based on reinforcement learning.

\myparagraph{Setup.}
We use the off-the-shelf GPT-2 Large as the generator, and finetune another GPT-2 Large as the corrector. 
During inference, we use nucleus sampling with $p=0.9$ to generate 25 samples for all baselines.
As the value function, we use the Perspective API score, $v(y) \in [0,1]$, which measures the toxicity of the completed sequence.
We do up to three corrections with the corrector model.

\myparagraph{Results.}
\begin{table}[t!]

\begin{minipage}{0.63\linewidth}
\setlength{\tabcolsep}{4pt}
    \centering\footnotesize
    \begin{tabular}{lcccccccc}
    \toprule
         & \multicolumn{2}{c}{\textbf{Toxicity}}& \textbf{Fluency}& \multicolumn{2}{c}{\textbf{Diversity}}\\
         \cmidrule(lr){2-3}\cmidrule(lr){4-4}\cmidrule(lr){5-6}
         & \textbf{Avg.~Max}. & \textbf{Prob.} & \textbf{Perplexity} & \textbf{dist-2} & \textbf{dist-3} \\
         \midrule
         GPT-2    & 0.527 & 0.520 & 11.31 & 0.85 & 0.85 \\
         \midrule
         PPLM~[\citenum{Dathathri2020PlugAP}]         & 0.520 & 0.518 & 32.58 & 0.86 & 0.86 \\
         GeDi~[\citenum{krause-etal-2021-gedi-generative}]          & 0.363 & 0.217 & 43.44 & 0.84 & 0.83 \\
         DExpert~[\citenum{liu-etal-2020-unsupervised}]     & 0.314 & 0.128 & 25.21 & 0.84 & 0.84 \\
         DAPT~[\citenum{gururangan-etal-2020-dont}]         & 0.428 & 0.360 & 31.22 & 0.84 & 0.84 \\
         PPO~[\citenum{quark22}]  & 0.218 & 0.044 & 14.27 & 0.79 & 0.82 \\
         Quark~[\citenum{quark22}]    & 0.196 & 0.035 & 12.47 & 0.80 & 0.84 \\
         \midrule
         \textsc{Self-Correct} & \textbf{0.171} & \textbf{0.026} & \textbf{11.81} & 0.80 & 0.83\\
    \bottomrule
    \end{tabular}
    \caption{\textbf{Toxicity reduction.} GPT-2 is the base generator.}
    \label{tab:toxicity_results}
    \end{minipage}
    \hfill
    \begin{minipage}{0.34\linewidth}
\centering 
\includegraphics[width=1.0\linewidth]{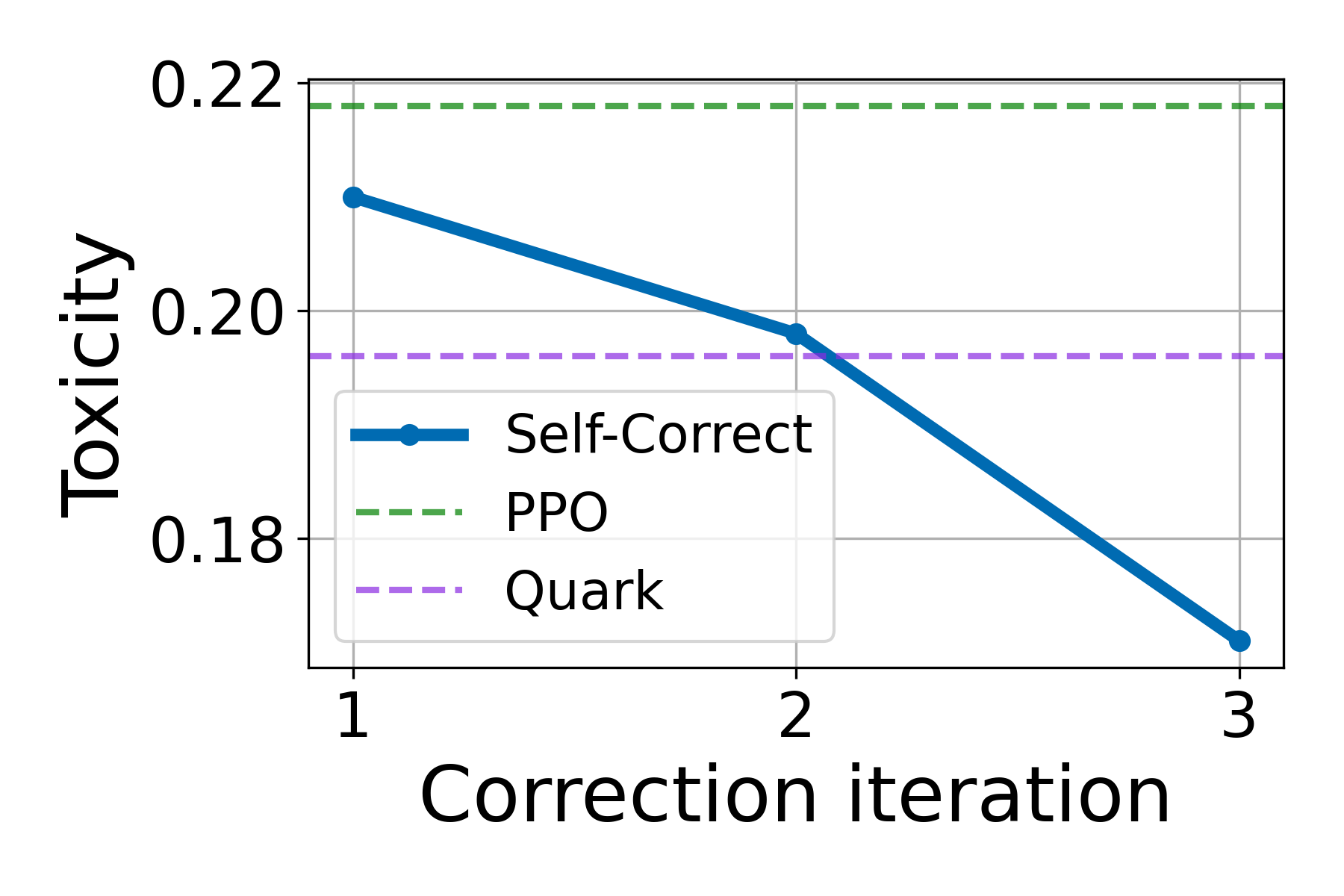}
\vspace{-1.5em}
\captionof{figure}{Applying multiple corrections reduces toxicity.}
\label{fig:toxicity}

\end{minipage}\hfill
\end{table}
\autoref{tab:toxicity_results} shows that \method reduces the rate of toxic generations substantially, while also maintaining fluency and diversity.
\method outperforms all baselines.
This includes inference-time algorithms (PPLM, GeDi, DExpert), which do not modify the generator but degrade fluency and yield higher toxicity compared to \method, as well as reinforcement learning methods (PPO, Quark) that adjust the generator using toxicity as a (negative) reward.
The results show that \method is  effective for detoxification, without having to modify the underlying generator. 
We study implications of this latter property further in the next section.

\subsection{Changing Modules -- Correcting GPT-3}
\label{sec:modularity}
Next, we show that a self-corrector can improve the outputs of a generator that is much larger than the corrector.
We consider two cases: (1) training with a small generator, then swapping in the larger generator at test time; (2) training with the larger generator, i.e. using the large generator to initialize the datapool for self-corrective learning, then using the large generator at test time.

\myparagraph{Toxicity.} We evaluate case (1) for reducing the toxicity of a large generator (GPT-2 XL, GPT-3).
We generate an initial sequence using the large generator, then refine it with our corrector trained in the previous experiments (\S\ref{ssec:toxicity}). \autoref{tab:modularity_combined} shows that
the resulting self-corrector (large generator + corrector) has substantially reduced toxicity compared to the large generator.
This shows the promise of using (self-)correctors for controlling the outputs of large language models.

\myparagraph{Math program synthesis.} \autoref{tab:modularity_combined} shows results for math.
Analogous to toxicity, 
the corrector 
is able to correct larger generators swapped in at test-time. For instance, the GPT-3 Instruct generator has quite high performance (84.90 Multitask, 36.80 GSM), which improves to 90.90 and 45.00, respectively, by adding in a corrector.
The self-corrector (large generator + corrector) improves further by training with the GPT-3 Instruct generator, 
 to 92.75 and 45.92, respectively.

\begin{table*}[t!]
    \centering\footnotesize
     \scalebox{.98}{
    \begin{tabular}{llllcccc}
    \toprule
     \textbf{Task} &    \textbf{Dataset} & \textbf{Generator (train)} & \textbf{Generator (test)} & \textbf{Generator} & \textbf{Self-corrector}\\
         \midrule
       \multirow{6}{*}{Math Synthesis $\uparrow$} & 
                            & Neo 1.3B & GPT-3 & 46.70 & 80.00\\
                          &  Multitask  & Neo 1.3B & GPT-3 Instruct & 84.90 & 90.90\\
                          &  & GPT-3 Instruct  & GPT-3 Instruct & 84.90 & 92.75\\
         \cmidrule{2-6}
        &  
                        & Neo 1.3B & GPT-3  & 6.96 & 24.30 \\
                       & GSM & Neo 1.3B & GPT-3 Instruct  & 36.80 & 45.00\\
                       & & GPT-3 Instruct  & GPT-3 Instruct  & 36.80 & 45.92\\
    \midrule
           \multirow{3}{*}{Detoxification \hspace{1.3mm}$\downarrow$} &   
                            & GPT2-L & GPT2-XL & 0.383 & 0.027\\
                          & RTPrompts & GPT2-L & GPT-3  & 0.182 & 0.025 \\
                          &  & GPT2-L  & GPT-3 Instruct &  0.275 & 0.023 \\
    \bottomrule
    \end{tabular}}
    \caption{\textbf{Modularity (program synthesis and detoxification).} Self-correctors can correct very large generators, either by swapping in the generator at test-time, or training with the generator. For math synthesis, the corrector is GPT-Neo 1.3B, and here we only correct incorrect outputs. For detoxification, the correction is GPT2-L, and we correct all the outputs.
    }
    \label{tab:modularity_combined}
\end{table*}

\subsection{Leveraging Explicit Feedback}
\label{sec:feedback}

Next, we demonstrate \method's capacity to incorporate explicit natural language feedback.
This amounts to defining a feedback function $f$, then using the same self-corrective learning and inference algorithms (\S\ref{ssec:learning}) as in our preceding experiments (in those experiments, $f$ returned $\emptyset$).
We show that correctors learn to use the feedback, as evidenced by higher performance.

\myparagraph{Toxicity.} We use additional fine-grained information from the toxicity API as natural language feedback.
 Specifically, besides the overall toxicity score, Perspective API also provides scores for fine-grained attributes of toxicity (e.g. identity attack, profanity, flirtation, etc.). 
 At training time, we compare the attribute scores from a hypothesis and its selected correction, and use the attribute with the largest decrease as natural language feedback (e.g. "decrease toxicity in \textit{profanity}").
 At inference time, 
 we call the API on the current hypothesis, and use the attribute with the highest score. 
 Here we use the API at inference time, which is \textit{not} required in our previous experiments.

\myparagraph{Lexical constraints.}
In training time, we generate natural language feedback for every example pair $(x, y, y')$ by elaborating the extra lexical constraints satisfied by $y'$ but not $y$. e.g. 
\textit{``adding constraint word: read''}.
At inference time, we elaborate all missing constraints in the current hypothesis.

\myparagraph{Math program synthesis.}
Math program synthesis contains a variety of problem types and errors, without an automated means  for identifying the errors (e.g. an API). 
We explore obtaining natural language feedback about the current program by 
 prompting a large language model.
 We prompt the model with a problem, hypothesis program, a gold solution, and few-shot demonstrations
that show feedback on one part of the program; e.g.
\textit{In the initial guess, 3 should be subtracted.}
When the program is correct, the feedback is \textit{Correct.}
At inference time, we also use feedback from the language model.
 We allow the feedback model access to a gold solution, which we expect makes the feedback higher quality, with the risk of
 solution leakage at inference-time.
 Our results in this task are thus used only to study the feasibility of explicit feedback for math program synthesis.

\begin{table*}[t!]
    \centering\footnotesize
    \begin{tabular}{lcccccccc}
    \toprule
         & \multicolumn{3}{c}{\textbf{Toxicity $\downarrow$}} & \multicolumn{2}{c}{\textbf{Constrained Gen. $\uparrow$}} & \multicolumn{2}{c}{\textbf{Math $\uparrow$}}\\
         \cmidrule(lr){2-4}\cmidrule(lr){5-6}\cmidrule(lr){7-8}
          & \textbf{Avg.~Max}. & \textbf{Prob.} & \textbf{Fluency} & \textbf{Fluency} & \textbf{Constraints} & \textbf{Correct} & \textbf{Correct$_*$}\\
         \midrule
         Generator & 0.527 & 0.520 & 11.31 & 14.97 & 91.38 & 49.02 & 49.02\\
         \method   & 0.171 & 0.026 & 11.81 & 15.30 & 94.58 & 74.31 & 79.80\\
         \ \ \textsc{+ feedback} &\textbf{0.156} & \textbf{0.020} & 11.86 & 15.24 & \textbf{95.88} & \textbf{81.76} & \textbf{82.35}\\
    \bottomrule
    \end{tabular}
    \caption{\textbf{Explicit natural language feedback.} Correct$_*$ means only correcting incorrect outputs.
    }
    \label{tab:feedback-all}
\end{table*}

\setlength{\columnsep}{0.2cm}
\begin{figure*}[t]
\setlength{\columnsep}{0.2cm}
\begin{multicols}{2}
\begin{tcolorbox}[colback=qualcolor!5!white,colframe=qualcolor!75!black]
\begin{small}
\textbf{Problem:}\\
Melanie had 19 dimes in her bank. Her dad gave her 39 dimes and her mother gave her 25 dimes. How many dimes does Melanie have now? 
\Sepline
\textbf{Generator (GPT-Neo):}
\vspace{-0.3em}
\begin{python}
answer = 19 + 25
print(answer)
\end{python}
\textbf{Feedback (GPT-3):}
\vspace{-0.3em}
\begin{python}
# In the initial guess, 
# 39 is not included.
\end{python}
\textbf{Corrector (GPT-Neo):}
\vspace{-0.3em}
\begin{python}
answer = 19 + 25 + 39
print(answer)
\end{python}
\end{small}
\end{tcolorbox}

\begin{tcolorbox}[colback=qualcolor!5!white,colframe=qualcolor!75!black]
\begin{small}
\textbf{Problem:}\\
Lana’s favorite band was holding a concert where tickets were 6 dollars each. Lana bought 8
tickets for herself and her friends and 2 extra tickets in case anyone else wanted to go. How much did
she spend?
\Sepline
\textbf{Generator (GPT-Neo):}
\vspace{-0.3em}
\begin{python}
answer=(6.0*8.0)
print(answer)
\end{python}
\textbf{Feedback (GPT-3):}
\vspace{-0.3em}
\begin{python}
# In the initial guess, 
# 2 tickets are not included.
\end{python}
\textbf{Corrector (GPT-Neo):}
\vspace{-0.5em}
\begin{python}
answer=(6.0*(8.0+2.0))
print(answer)
\end{python}
\end{small}
\vspace{-0.9em}
\end{tcolorbox}
\end{multicols}
\vspace{-1.7em}
\caption{\textbf{Self-correction with natural language feedback.}
}
\label{fig:math-feedback-examples}
\end{figure*}

\myparagraph{Setup.} For toxicity, lexical constraints, and math we use \textsc{RealToxicityPrompts}, \textsc{CommonGen}, and the \textsc{Multitask} arithmetic setting, respectively. We follow the setup of each task's previous experiments (\S\ref{ssec:toxicity},\S\ref{ssec:constrained},\S\ref{ssec:math}), except for math we use 5 correction iterations (previously 1).
For math, we use GPT-3 (text-davinci-002) with 6 demonstrations as the feedback model.

\myparagraph{Results.}
\autoref{tab:feedback-all} shows that explicit natural language feedback improves performance in all three tasks.
For toxicity, this means that providing fine-grained attributes (e.g. identity attack, profanity, etc.) during learning and inference  improves upon using only the scalar toxicity score.
Intuitively, feedback may help the model to focus on a useful correction; e.g., see \autoref{fig:math-feedback-examples}.

\subsection{Additional Ablations and Analysis}
\label{subsec:ablation}

\myparagraph{Effect of multiple corrections.}
Previously, Figure~\ref{fig:toxicity} showed that multiple corrections led to better toxicity reduction.
On math (Multitask setting), Figure~\ref{fig:math-feedback} shows that performance improves with more than one correction, and that
multiple corrections are more beneficial with feedback.
Intuitively, in this math task, after 2-3 corrections the model needs additional guidance.

\myparagraph{Effect of pairing and proportional sampling.}
Self-corrective learning (i) samples pairs for learning proportional to \autoref{eqn:residual-subsample}, (ii) only pairs sequences that improve value.
We ablate these features by training on Multitask using a  data pool that samples a pair for learning uniformly (rather than  \autoref{eqn:residual-subsample}), and a data pool without value pairing.
\autoref{tab:pairing} shows that both improve performance. 

\myparagraph{Effect of exploration.}
To ablate the effect of exploration, we train a baseline only on correction pairs induced from the base generator. 
\autoref{tab:multiple-corrections}
 shows results on the three math datasets, indicating that exploration improves performance. 

\begin{table}[t!]
\begin{minipage}{0.44\linewidth}
\centering 
\includegraphics[width=0.9\linewidth]{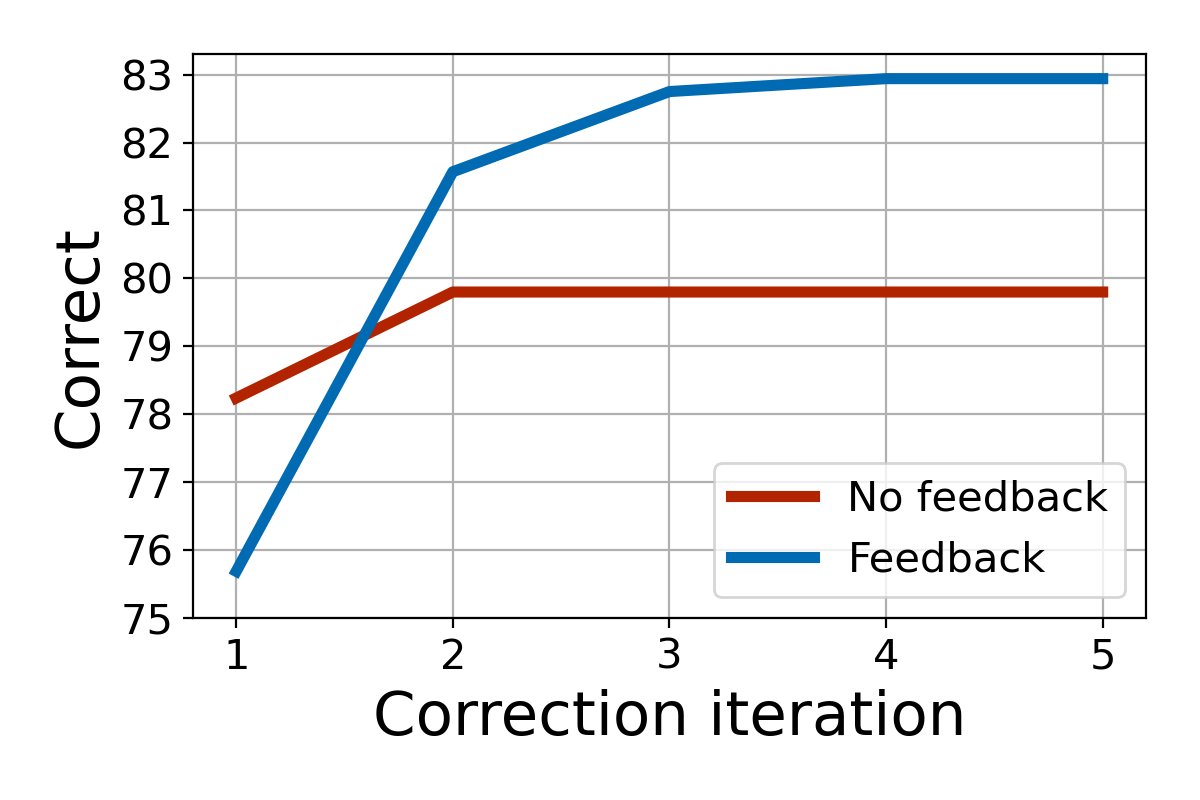}
\vspace{-1em}
\captionof{figure}{Math: multiple corrections.}
\label{fig:math-feedback}

\end{minipage}\hfill
\begin{minipage}{0.54\linewidth}
\centering\footnotesize
 \scalebox{.98}{
    \begin{tabular}{lccc}
    \toprule
    \textbf{Ablation} & \multicolumn{1}{c}{\textbf{Math}}& \textbf{\textsc{CommonGen}} \\
    \midrule
    \method & \textbf{78.24} & \textbf{94.55}\\
     \ \ \xmark\ proportional sampling &  77.25 & 93.49\\
     \ \ \xmark\ value pairing &  62.35 & 91.76 \\
    \bottomrule
    \end{tabular}}
    \vspace{-0.5em}
    \caption{Effect of pairing and proportional sampling.}
    \label{tab:pairing}
\begin{tabular}{ccccccccc}
    \toprule
    \textbf{Exploration} & \textbf{Multiarith} & \textbf{Multitask} & \textbf{GSM8k}\\
    \midrule
    \xmark & 89.20 & 73.49 & 17.60 \\
    \cmark & \textbf{99.17} & \textbf{78.24} & \textbf{23.96}\\
    \bottomrule
    \end{tabular}
    \vspace{-0.5em}
    \caption{Effect of exploration on program synthesis.}
    \label{tab:multiple-corrections}
\end{minipage}
\end{table}

\section{Related Work}
\vspace{-0.5em}
Self-correction relates to recent works on editing text, including modeling Wikipedia edits \citep{reid2022learning,faltings-etal-2021-text,schick2022peer}, which relies on supervised edits, unsupervised methods \citep{Miao_Zhou_Mou_Yan_Li_2019,liu-etal-2020-unsupervised} that perturb sequences with simple operations (e.g. insertion, deletion),
 editing with models trained on human-written critiques~\citep{saunders2022self}, or iteratively updating continuous variables~\citep{lee-etal-2020-iterative,Li2022DiffusionLMIC,qin2022cold}.
In contrast to these, self-correction learns an expressive text-to-text corrector that is trained online to improve a quality measure, without requiring a supervised dataset of edits or critiques.
Separately, denoising ground-truth sequences is a common pretraining objective~\citep{devlin-etal-2019-bert,lewis-etal-2020-bart,RaffelT5}, while self-correction `denoises' generations to improve a scalar quality measure.
Scalar measures are often improved with reinforcement learning (RL) on a base generator~\citep{ziegler2019finetuning,stiennon2020,quark22}, which is infeasible for improving many language models (e.g. those accessed through an API), and uses only scalar feedback. 
Moreover, self-correction learns the difference between a generation and solution, and is  complementary to RL-tuned generators, which can be used within a self-corrector.
Finally, self-correction decomposes  generation into multiple steps, which relates to methods 
that generate rationales before a response~\citep{Wei2022ChainOT,Dohan2022LanguageMC}.
Self-correction also produces  intermediate steps, but each step is of the same form as the output, allowing for re-using previous generations.

\section{Conclusion}
\vspace{-0.5em}
We introduced self-correctors, a class of models that decompose generation into initial generation and correction steps.
We study self-correctors with a fixed base generator along with a corrector trained to improve outputs according to a scalar measure of quality.
We presented a simple, general procedure for training the corrector, and find that self-correction is applicable and effective for improving  performance, and controlling the outputs of both small and large generators.
Moreover, we found that self-correction along with our learning framework provides a promising mechanism for using natural language feedback to improve generation.
These findings, along with exploring alternative self-correctors, open up many avenues that we leave for future work.

\section*{Acknowledgments}
This work was funded in part by the DARPA MCS program through NIWC Pacific (N66001-19-2-4031), and the Allen Institute for AI.

\bibliography{iclr2023_conference}
\bibliographystyle{iclr2023_conference}

\clearpage
\newpage
\appendix

\begin{center}
{\LARGE \textsc{
Appendix
}}
\vspace{40pt}
\end{center}

\section{Additional Experimental Details}
\subsection{Mathematical Program Synthesis}
We fine-tune a separate instance of GPT-Neo 1.3B as an initial generator, using the Huggingface library with default hyperparameters, except for evaluation steps, which we set to a small number to ensure a strong checkpoint is selected for each dataset.
We use the fine-tuned initial generator as initialization for the corrector,
and tune the corrector on sequences
${\small
\texttt{[SC]x[CURR]yi[START]yj[END]},
}$
where $x$ is a problem, $y_i$ and $y_j$ form a residual pair, and $[\cdot]$ are special tokens.
The loss is on tokens after $\texttt{[START]}$.

\paragraph{Feedback.}
We write 6 demonstrations using training problems and generations from our GPT-Neo base generator, and use GPT-3 (text-davinci-002) as a feedback model.
We use the same training procedure and hyperparameters, except that the sequences now include feedback,
${\small
\texttt{[SC]x[CURR]yi[FEEDBACK]F(x,yi)[START]yj[END]},
}$
where $x$ is a problem, $y_i$ and $y_j$ form a residual pair, and $F(x,y_i)$ is feedback.
We include loss on tokens after $\texttt{[FEEDBACK]}$.

\subsection{Lexically-constrained Generation}

\textbf{Hyper-parameters.  }\autoref{tab:hyper-cg} and \autoref{tab:hyper-e2e} show hyperparameters for CommonGen and E2E.

\textbf{Human Evaluation. } We evaluate fluency of generations in E2E task using human annotators on Amazon Mechanical Turk (AMT). We randomly sampled 100 instances, along with generations of different baselines and self-corrections. For each instance, we ask 3 annotators to evaluate the fluency of generations on a 3-point Likert scale. We aggregate annotations from 3 annotators using majority vote. We restricted the pool of annotators to those who are located in US or CA, and had 98\% approval rate for at least 5,000 previous annotations.
\begin{table}[h]
\begin{minipage}{0.48\linewidth}
\centering 
    \centering\footnotesize
    \begin{tabular}{lc}
    \toprule
    \textbf{Hyperparameter} & \textbf{Assignment}\\
    \midrule
    Predictor & GPT-2$_{Large}$ \\
    \# steps & 6000\\
    batch size & 128\\
    optimizer & Adam\\
    learning rate & $1.e^-5$ \\
    decoding alg. & beam search (k=5) \\
    \bottomrule
    \end{tabular}
    \caption{
Hyperparameters for \textsc{CommonGen}.
    }
    \label{tab:hyper-cg}
\end{minipage}\hfill
\begin{minipage}{0.5\linewidth}
\centering\footnotesize
    \begin{tabular}{lc}
    \toprule
    \textbf{Hyperparameter} & \textbf{Assignment}\\
    \midrule
    Predictor & GPT-2$_{Medium}$  \\
    \# steps & 10000\\
    batch size & 100 \\
    optimizer & Adam\\
    learning rate & $1.e^-5$ \\
    decoding alg. & beam search (k=5) \\
    \bottomrule
    \end{tabular}
    \caption{Hyperparameters for E2E.
    }
    \label{tab:hyper-e2e}
\end{minipage}
\end{table}

\section{Additional Results}


\begin{table*}[h]
    \centering\footnotesize
    \begin{tabular}{lcccccccc}
    \toprule
         & \multicolumn{2}{c}{\textbf{Toxicity}}& \textbf{Fluency}& \multicolumn{2}{c}{\textbf{Diversity}}\\
         \cmidrule(lr){2-3}\cmidrule(lr){4-4}\cmidrule(lr){5-6}
         & \textbf{Avg.~Max}. & \textbf{Prob.} & \textbf{Perplexity} & \textbf{dist-2} & \textbf{dist-3} \\
         \midrule
         GPT2-L        & 0.527 & 0.520 & 11.31 & 0.85 & 0.85 \\
         \textsc{Self-Correct} & 0.171 & 0.026 & 11.81 & 0.80 & 0.83\\
         \textsc{Self-Correct + feedback} & \textbf{0.156} & \textbf{0.020} & 11.86 & 0.80 & 0.83\\
    \bottomrule
    \end{tabular}
    \caption{Evaluation results of toxicity reduction experiments with  natural language feedback.}
    \label{tab:toxicity_feedback}
\end{table*}

\begin{table*}[h]
    \centering\footnotesize
    \begin{tabular}{lcccccccc}
    \toprule
         & \textbf{Bleu-4} & \textbf{CIDER} & \textbf{Coverage} & \textbf{Runtime}\\
         \midrule
         NeuroLogic~[\citenum{lu-etal-2021-neurologic}]          & 26.70 & 14.70 & 97.70& 2.04s/sent\\
         NeuroLogic-A*esque~[\citenum{lu2022neurologicastar}]       & 28.20 & 15.20 & 97.80   & 19.24s/sent\\
         \midrule
         GPT-2        & 27.90 & 14.97 & 91.38 & 0.2s/sent\\
         \textsc{Self-Correct}  & 27.98 & 15.30 & 94.58 & 0.8s/sent\\
          \textsc{Self-Correct} + feedback & 27.82 & 15.24 & 95.88 & 0.8s/sent\\
         \textsc{Self-Correct}+NeuroLogic  & 28.17 & 15.28 & \textbf{97.80} & 2.24s/sent\\
    \bottomrule
    \end{tabular}
    \caption{Evaluation rresults of lexically-constrained generation on \textsc{CommonGen}.
    }
    \label{tab:commongen_results}
\end{table*}

\begin{table*}[h]
\setlength\tabcolsep{2.5pt}
    \centering\footnotesize
    \begin{tabular}{lcccccccc}
    \toprule
         & \textbf{Coverage} &  \textbf{BLEU-4} & \textbf{NIST} & \textbf{R-L}  & \textbf{METEOR} & \textbf{CIDER} \\
         \midrule
        \textsc{Prefix-Tuning}~\citep{li-liang-2021-prefix} & 91.16 & 70.30 & 8.82 & 72.10 & 46.30 & 2.46 \\
         \midrule
         GPT-2    & 91.50 & 67.12 & 8.67 & 70.25  & 45.58 & 2.33\\
         \textsc{Self-Correct} & \textbf{98.77} & 68.81  & 8.78 & 68.60 & 45.11 & 2.38  \\
    \bottomrule
    \end{tabular}
    \caption{Evaluation results of lexically-constrained generation on E2E. 
    }
    \label{tab:e2e_results}
\end{table*}

\section{Qualitative Examples}

\setlength{\columnsep}{0.2cm}
\begin{figure}[t]

\setlength{\columnsep}{0.2cm}
\begin{multicols}{2}
\begin{tcolorbox}[colback=qualcolor!5!white,colframe=qualcolor!75!black]
\begin{small}
\textbf{Problem:}\\
Ralph watches TV for 4 hours a day from Monday to Friday, and 6 hours a day on Saturday and Sunday. How many hours does Ralph spend watching TV in one week?
\Sepline
\vspace{-1.2em}
\begin{multicols}{2}
\textbf{Generator:}
\vspace{-0.3em}
\begin{python}
a=4*7
b=6*7
c=a+b
answer=c
print(answer)
\end{python}
\columnbreak

\textbf{Corrector:}
\vspace{-0.3em}
\begin{python}
a=4*5
b=6*2
c=a+b
answer=c
print(answer)
\end{python}
\end{multicols}
\end{small}
\end{tcolorbox}

\begin{tcolorbox}[colback=qualcolor!5!white,colframe=qualcolor!75!black]
\begin{small}
\textbf{Problem:}\\
The pirates plan to explore 4 islands. Two islands require walking 20 miles per day while the other two islands require 25 miles per day. How many miles will they have to walk if it takes 1.5 days to explore each island?
\Sepline
\vspace{-1.2em}
\begin{multicols}{2}
\textbf{Generator:}
\vspace{-0.3em}
\begin{python}
a=20*2
b=25*2
c=a+b
d=c*1.5
e=d+b
answer=e
print(answer)
\end{python}
\columnbreak
\textbf{Corrector:}
\vspace{-0.5em}
\begin{python}
a=20*2
b=25*2
c=a+b
d=c*1.5
answer=d
print(answer)
\end{python}

\end{multicols}
\end{small}
\end{tcolorbox}
\end{multicols}
\vspace{-1.7em}
\caption{\textbf{Grade school math (GSM) self-corrections.} Left: the structure of the generator's solution is valid, but it incorrectly uses the total number of days in a week for both \textit{a} and \textit{b}; the corrector fixes the program to correctly account for the 5 weekdays and 2 weekend days. Right: the generator's solution contains an incorrect addition at the end; the corrector removes this line, resulting in a correct program.
}
\label{fig:math-examples2}

\begin{multicols}{2}
\begin{tcolorbox}[colback=qualcolor!5!white,colframe=qualcolor!75!black]
\begin{small}
\textbf{Problem:}\\
A spiral notebook costs 15, and a personal planner costs \$10. How much would it cost in total to buy 4 spiral notebooks and 8 personal planners at a 20\% discount?
\Sepline
\vspace{-1.2em}
\begin{multicols}{2}
\textbf{Generator:}
\vspace{-0.3em}
\begin{python}
a=4*15
b=8*10
c=a+b
answer=c
print(answer)
\end{python}
\columnbreak

\textbf{Corrector:}
\vspace{-0.3em}
\begin{python}
a=4*15
b=8*10
c=a+b
d=c*20/100
e=c-d
answer=e
print(answer)
\end{python}
\end{multicols}
\end{small}
\end{tcolorbox}

\begin{tcolorbox}[colback=qualcolor!5!white,colframe=qualcolor!75!black]
\begin{small}
\textbf{Problem:}\\
Julia has \$40. She spends half of her money to buy a new game for her phone. She spends a quarter of what she has left on in-game purchases. How much money does she have left?
\Sepline
\vspace{-1.2em}
\begin{multicols}{2}
\textbf{Generator:}
\vspace{-0.3em}
\begin{python}
a=40/2
b=40-a
c=b/4  
d=b-c
e=d+40 
answer=e 
print(answer)
\end{python}
\columnbreak
\textbf{Corrector:}
\vspace{-1.5em}
\begin{python}
a=40/2
b=a/4
c=a-b
answer= c
print(answer)
\end{python}

\end{multicols}
\end{small}
\end{tcolorbox}
\end{multicols}
\vspace{-1.7em}
\caption{\textbf{Grade school math (GSM) self-corrections.} Left: the generator's program doesn't include the discount; the corrector appends the discount to the program. Right: a more sophisticated multipart correction. The generator's assignment of \textit{b} (line 2), and addition to \textit{e} (line 5) are incorrect. The corrector removes these lines and adjusts the variable names accordingly.
}
\label{fig:math-examples3}
\end{figure}

\setlength{\columnsep}{0.2cm}
\begin{figure}[t]
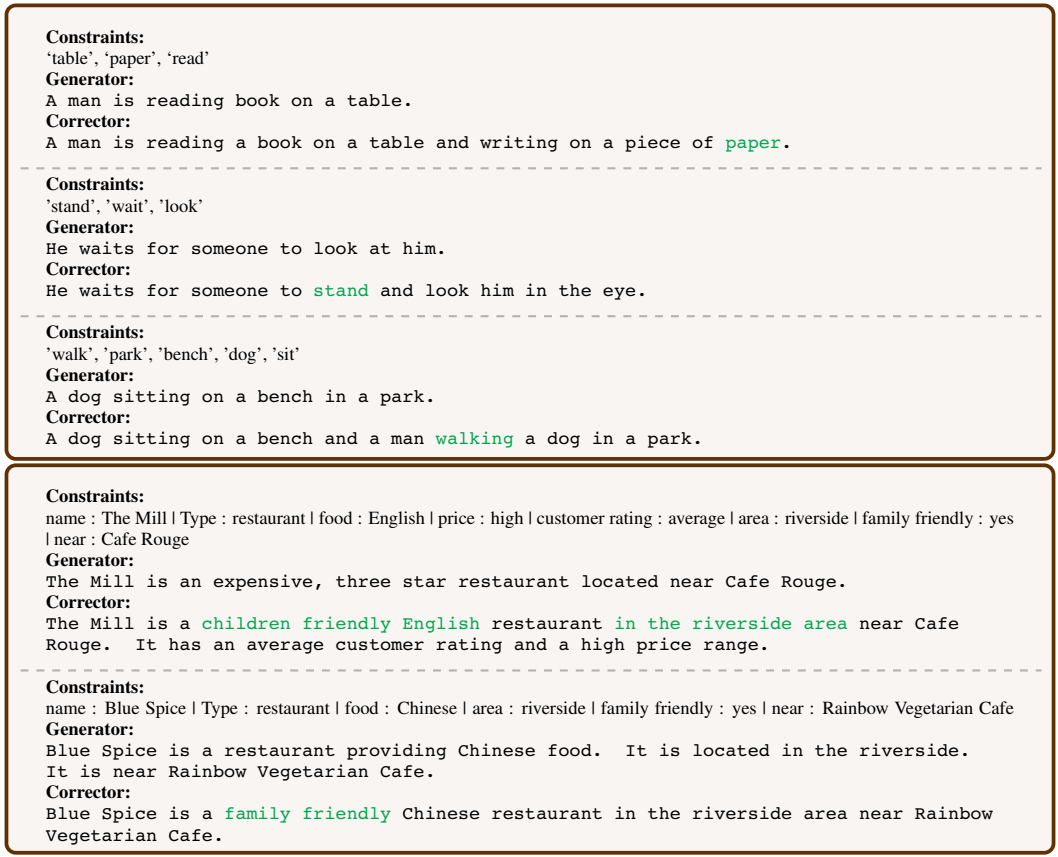


\begin{tcolorbox}[colback=qualcolor!5!white,colframe=qualcolor!75!black]
\begin{scriptsize}
\textbf{Constraints:}\\
`table', `paper',  `read' \\
\textbf{Generator:} \\
\texttt{A man is reading book on a table.} \\
\textbf{Corrector:} \\
\texttt{A man is reading a book on a table and writing on a piece of \textcolor{green(pigment)}{paper}. }
\Sepline

\textbf{Constraints:} \\'stand', 'wait', 'look'\\
\textbf{Generator:}\\
\texttt{He waits for someone to look at him.}\\
\textbf{Corrector:}\\
\texttt{He waits for someone to \textcolor{green(pigment)}{stand} and look him in the eye.}
\Sepline

\textbf{Constraints:} \\'walk', 'park', 'bench', 'dog', 'sit'\\
\textbf{Generator:}\\
\texttt{A dog sitting on a bench in a park.}\\
\textbf{Corrector:}\\
\texttt{A dog sitting on a bench and a man \textcolor{green(pigment)}{walking} a dog in a park.}

\end{scriptsize}
\vspace{-0.5em}
\end{tcolorbox}

\begin{tcolorbox}[colback=qualcolor!5!white,colframe=qualcolor!75!black]
\begin{scriptsize}

\textbf{Constraints:}\\
name : The Mill | Type : restaurant | food : English | price : high | customer rating : average | area : riverside | family friendly : yes | near : Cafe Rouge \\
\textbf{Generator:} \\
\texttt{The Mill is an expensive, three star restaurant located near Cafe Rouge.} \\
\textbf{Corrector:} \\
\texttt{The Mill is a \textcolor{green(pigment)}{children friendly} \textcolor{green(pigment)}{English} restaurant \textcolor{green(pigment)}{in the riverside area} near Cafe Rouge. It has an average customer rating and a high price range.}

\Sepline

\textbf{Constraints:}\\
name : Blue Spice | Type : restaurant | food : Chinese | area : riverside | family friendly : yes | near : Rainbow Vegetarian Cafe
\textbf{Generator:} \\
\texttt{Blue Spice is a restaurant providing Chinese food. It is located in the riverside. It is near Rainbow Vegetarian Cafe.} \\
\textbf{Corrector:} \\
\texttt{Blue Spice is a \textcolor{green(pigment)}{family friendly} Chinese restaurant in the riverside area near Rainbow Vegetarian Cafe.}

\end{scriptsize}
\vspace{-0.6em}
\end{tcolorbox}

\caption{\textbf{Constrained generation self-corrections.} Top: \textsc{CommonGen}, and Bottom: E2E. Words in \textcolor{green(pigment)}{green} indicate constraints that were previously missing.
}
\label{fig:examples-cg-e2e}
\end{figure}

\setlength{\columnsep}{0.2cm}
\begin{figure}[t]
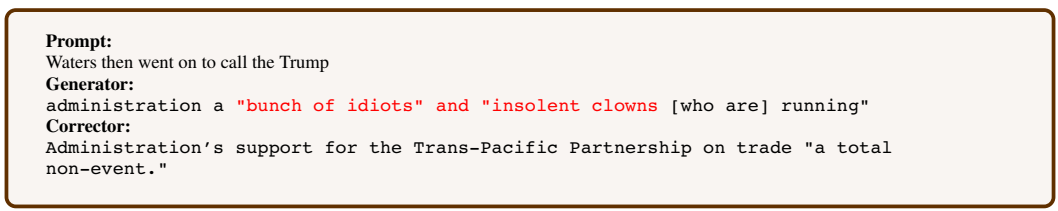


\begin{tcolorbox}[colback=qualcolor!5!white,colframe=qualcolor!75!black]
\begin{scriptsize}
\textbf{Prompt:}\\
Waters then went on to call the Trump \\
\textbf{Generator:} \\
\texttt{administration a \textcolor{red}{"bunch of idiots" and "insolent clowns} [who are] running"} \\
\textbf{Corrector:} \\
\texttt{Administration's support for the Trans-Pacific Partnership on trade "a total non-event."} \\
\texttt{}

\end{scriptsize}
\vspace{-0.5em}
\end{tcolorbox}

\caption{\textbf{Toxicity control self-corrections.} Words in \textcolor{red}{red} indicate toxic language.
}
\label{fig:examples-toxicity}
\end{figure}

\end{document}